\documentclass{article}

\PassOptionsToPackage{numbers, compress}{natbib}

\ifx\planningworkshop\undefined
\usepackage[final]{aloe_2023_neurips}
\else
\usepackage[final]{neurips_2023}
\fi

\usepackage[utf8]{inputenc} 
\usepackage[T1]{fontenc}    
\usepackage{hyperref}       
\usepackage{url}            
\usepackage{booktabs}       
\usepackage{amsfonts}       
\usepackage{nicefrac}       
\usepackage{microtype}      
\usepackage{xcolor}         
\usepackage{subcaption}
\usepackage{caption}

\title{Mini-BEHAVIOR: A Procedurally Generated Benchmark for Long-horizon Decision-Making in Embodied AI}

\usepackage{subfiles} 
\usepackage{graphicx}

\author{%
Emily Jin$^{1} \thanks{Equal Contribution}$ \quad Jiaheng Hu$^{2} \footnotemark[1]$ \quad Zhuoyi Huang $^1$ \\ 
\textbf{Ruohan Zhang}$^{1}$ \quad \textbf{Jiajun Wu}$^1$ \quad \textbf{Li Fei-Fei}$^1$  \quad \textbf{Roberto Mart{\'i}n-Mart{\'i}n}$^{2}$\\
$^1$Stanford University \quad $^2$University of Texas at Austin \\
\texttt{\{emilyjin,zhuoyih,zharu\}@stanford.edu}\\
\texttt{\{jhu, robertomm\}@cs.utexas.edu}\\
\texttt{\{jiajunwu, feifeili\}@cs.stanford.edu}
}

\begin{document}
\maketitle

\ifx\planningworkshop\undefined

\begin{abstract}

We present Mini-BEHAVIOR, a novel benchmark for embodied AI that challenges agents to use reasoning and decision-making skills to solve complex activities that resemble everyday human challenges. The Mini-BEHAVIOR environment is a fast, realistic Gridworld environment that offers the benefits of rapid prototyping and ease of use while preserving a symbolic level of physical realism and complexity found in complex embodied AI benchmarks. We introduce key features such as procedural generation, to enable the creation of countless task variations and support open-ended learning. Mini-BEHAVIOR provides implementations of various household tasks from the original BEHAVIOR benchmark, along with starter code for data collection and reinforcement learning agent training. In essence, Mini-BEHAVIOR offers a fast, open-ended benchmark for evaluating decision-making and planning solutions in embodied AI. It serves as a user-friendly entry point for research and facilitates the evaluation and development of solutions, simplifying their assessment and development while advancing the field of embodied AI. 
Code is publicly available at \url{https://github.com/StanfordVL/mini_behavior}.
\end{abstract}

\else
\begin{abstract}
We present Mini-BEHAVIOR, a novel benchmark for embodied AI that challenges agents to plan and solve complex activities resembling everyday human household tasks. The Mini-BEHAVIOR environment extends the widely used MiniGrid grid world with new modes of actuation, combining navigation and manipulation actions, multiple objects, states, scenes, and activities defined in first-order logic. 
Mini-BEHAVIOR implements various household tasks from the original BEHAVIOR benchmark, along with starter code for data collection and reinforcement learning agent training.
Together with Mini-BEHAVIOR, we also include a procedural generation mechanism to create countless variations of each task and support the study of plan generalization and open-ended learning.
Mini-BEHAVIOR is fast and easy to use and extend, providing the benefits of rapid prototyping while striking a good balance between symbolic-level decision-making and physical realism, complexity, and embodiment constraints found in complex embodied AI benchmarks. 
Our goal with Mini-BEHAVIOR is to provide the community with a fast, easy-to-use and modify, open-ended benchmark for developing and evaluating decision-making and generalizing planning solutions for embodied AI. 
Code is available at \url{https://github.com/StanfordVL/mini_behavior}.

\end{abstract} 
\fi

\section{Introduction}

Embodied AI focuses on developing agents that can interact with the environment and perform complex tasks. As in other AI fields, progress in embodied AI has largely been driven and supported by a rich set of benchmarks that enable researchers to focus on common challenges and measure progress in equal and fair conditions ~\citep{rearrangement, tdw_transport,virtualhome, alfred, igibson, metaworld}. 
The impressive progress in the field has led to increasingly richer and more complex benchmarks, pushing the boundaries of AI agents to perform navigation ~\cite{zhu2017target, zhu2021deep}, stationary manipulation \cite{rajeswaran2017learning, matas2018sim}, and their combination in mobile manipulation ~\cite{hu2023causal, li2020hrl4in}.
However, these more complex benchmarks are costly to run and train agents for, leading to a high entry point and slow development.

A complex benchmark with those characteristics is BEHAVIOR, \emph{a benchmark for everyday household activities in virtual, interactive, and ecological environments}. BEHAVIOR (both the 100 and 1K variants) seeks to drive the development of embodied agents by challenging them to perform everyday household tasks that are diverse, realistic, and long horizon~\citep{behavior-1k,behavior-100}. With the support of advanced physics simulators~\citep{igibson} and several interactive simulated scenes, BEHAVIOR defines multiple and diverse household tasks, ranging from \texttt{installing a printer} to \texttt{cleaning up the kitchen} to \texttt{preparing a salad}. These tasks are long-horizon and heterogeneous: they require thousands of decision-making steps to be completed and for the agent to master a variety of skills, from pick and place, to open/close, to switch on/off. 
A significant challenge in BEHAVIOR arises from 
\ifx\planningworkshop\undefined
the need to generalize to all possible variations of houses the AI agent may encounter, which requires open-ended training in different scenes.
\else
the need to generalize to reason and plan tasks combining multiple objects, actions, and state changes. 
\fi


To facilitate the process of creating solutions for embodied AI, researchers and AI practitioners developed simplified versions of these complex benchmarks~\citep{griddly, mazebase, overcooked}. These environments are simple, lightweight, and fast to run, supporting primitive decision-making tasks. One of the most used ones is MiniGrid~\citep{MiniGridMiniworld23}, with simplified versions of navigation. In MiniGrid, dynamics, actions, and states are discretized, simplifying simulation and decision-making. Due to its simplicity, it is highly reconfigurable and easy to extend, becoming one of the most used tools by embodied AI practitioners to create, test, and develop algorithmic ideas.
However, MiniGrid and the other reduced AI benchmarks fail to provide a simpler developing version of complex benchmarks for long-horizon, heterogeneous AI tasks such as BEHAVIOR, combining navigation and manipulation, with richer semantics, involving complex task planning, and with practical applications such as household activities.

We present a new simplified benchmark, \textbf{Mini-BEHAVIOR}, that provides the benefits of simple benchmarks for fast prototyping and training, while keeping the most relevant problem structure and the characteristics from the task-level decision-making challenge in BEHAVIOR: the long-horizon, heterogenousness of the activities, and the multi-object, multi-state, high variability of household environments.
Mini-BEHAVIOR re-defines BEHAVIOR tasks in a novel 3D Gridworld environment with support for procedural generation for potentially infinite variants of each task. 

We extend over previous gridworld environments by including semantically classified objects --\texttt{printer}, \texttt{plate}, \texttt{apple},\dots-- with multiple possible states --\texttt{frozen}, \texttt{onTop}, \texttt{closed},\dots-- in 2D gridworlds with rooms and furniture where we added a third dimension to enable a simplified vertical axis --\texttt{plates} can be \texttt{onTop} a \texttt{table}. We also extended the agents' capabilities to be able to change those objects' states and observe them.
We provide implementation for multiple household activities from the original BEHAVIOR, and starter code to collect data and train reinforcement learning agents, while keeping the simplicity and speed of the MiniGrid benchmark.



\begin{figure}[t!]
    \centering \includegraphics[width=\textwidth]{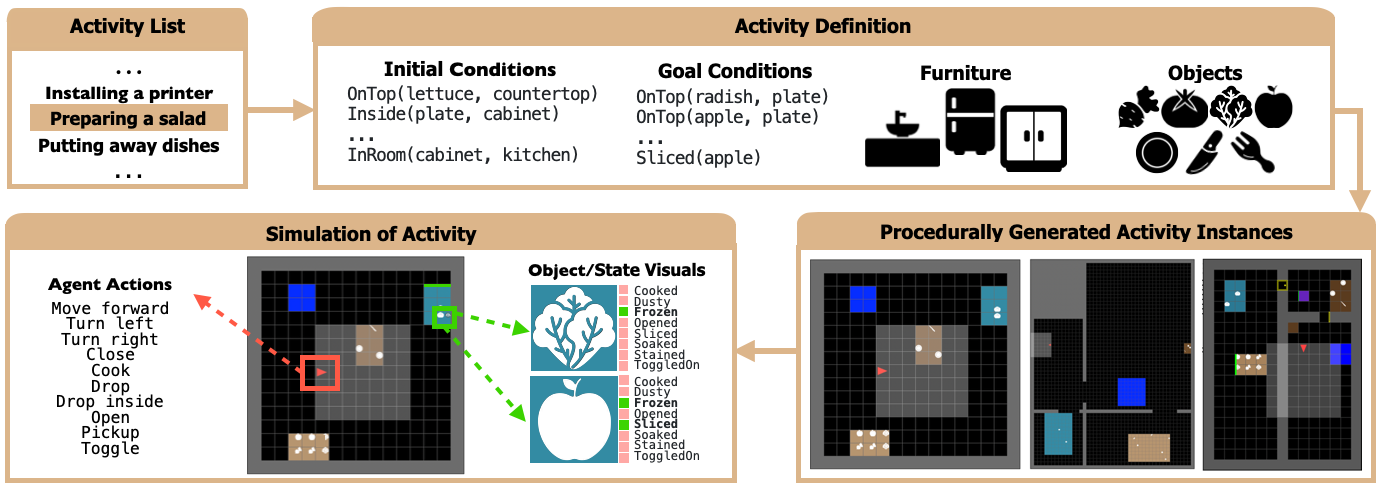}
    \caption{From Activity Definition to Simulated Activity Instance in Mini-BEHAVIOR: Given a long-horizon household activity definition in first-order logic, together with a dataset of simple object and furniture models, our Mini-BEHAVIOR benchmark procedurally generates unlimited activity instances in a fast, easy-to-use 3D Gridworld environment.}
    \label{fig:overall}
\end{figure}

In summary, the Mini-BEHAVIOR benchmark contains multiple long-horizon heterogenous tasks, is simple and fast, and open-ended through procedural generation. We hope that with Mini-BEHAVIOR, we provide the community with:
\begin{enumerate}
    \item a simple tool to standardize the evaluation of task-level decision-making and planning solutions, going beyond pure symbolic domains into problems with the constraints of embodied AI;
    \item an open-ended generative system that can create potentially infinite variations of each task;
    \item an easy starting point and fast prototyping platform for embodied AI solutions, keeping enough similarities to its more realistic, fully simulated version that we hope it will enable the easy transfer of some of the solutions;
\end{enumerate}

\section{Related Work}
Datasets and benchmarks have long played an important role in driving progress in AI. In embodied AI, several benchmarks have been developed to calibrate progress in the field, including Rearrangement, TDW Transport Challenge, VirtualHome, ALFRED, Interactive Gibson Benchmark, MetaWorld, and BEHAVIOR~\citep{rearrangement, tdw_transport,virtualhome, alfred, igibson, metaworld}. Of these, BEHAVIOR is the most comprehensive benchmark, with realistic, diverse, and complex tasks. BEHAVIOR tasks include household activities that are representative of the kinds of activities that humans face in their everyday lives. BEHAVIOR activities are also diverse in the sense that they involve a variety of scenes, objects, and actions, and they are complex since they require both low-level manipulation and high-level planning skills. 

One key feature of BEHAVIOR and other embodied AI benchmarks is that they use complex simulator environments to physically represent the real world. While this physical realism makes these benchmarks highly valuable to the robot learning community, they can incur huge computational costs and drastically reduce the prototyping speed, making them unsuitable for rapid experimentation and development.

To address the need for more accessible benchmarking tools, researchers have created simplified versions of these benchmarks in 2D-Gridworld environments to benchmark decision-making algorithms. Environments such as Griddly, MazeBase, Overcooked, and MiniGrid were developed to benchmark RL agents for 2D-games and goal-oriented tasks~\citep{griddly, mazebase, overcooked, MiniGridMiniworld23}. Thus far, MiniGrid is the best for supporting various goal-oriented tasks, and is known to be fast and easily customizable.

Even so, MiniGrid is only able to support a small range of goal-oriented tasks and lacks the complexity to support realistic activities such as the ones in BEHAVIOR. It provides few actions for interacting with the environment, limited objects, and physical grid constraints. Each cell is only allowed to have one object, and each object takes up exactly one cell. As a result, MiniGrid is 2D, and there are no notions of size, height, and vertical or internal placement. For example, both an apple and a bed have the same effective size and height, and it is not possible to place a pan on top of a counter or an apple inside the refrigerator. 

By comparison, Mini-BEHAVIOR extends MiniGrid's capabilities by adding procedural generation capabilities, as well as support for symbolic states, more objects and primitive actions, a third vertical dimension, and multi-cell objects. In the meantime, Mini-BEHAVIOR is built on top of the MiniGrid library and inherits aspects such as its speed and simplicity \cite{MiniGridMiniworld23}

\section{Mini-BEHAVIOR Benchmark Features}

With Mini-BEHAVIOR, we hope to bridge the gap between the complexities of benchmarks like BEHAVIOR and the simplicity of MiniGrid. 
The Mini-BEHAVIOR benchmark includes 20 household tasks of varying difficulty. The tasks are instantiated in a simulated household environment with support for 96 objects, 23 states, and 15 actions. The three key features of the benchmark are:
\begin{enumerate}
    \item \textbf{diverse and complex tasks:} a standardized set of 20 tasks for reproducible research that require reasoning and high-level planning skills;
    \item \textbf{fast simulation environment:} a simple, lightweight, fast, and easy-to-use Gridworld environment, suitable for rapid prototyping of decision-making algorithms;
    \item \textbf{procedural generation:} ability to generate unlimited activity instances with physically realistic object placement to support open-ended learning
    
\end{enumerate}


We proceed by discussing the three key features in greater detail. 

\label{s:intro}

\begin{figure}[t!]
    \centering \includegraphics[width=0.9\textwidth]{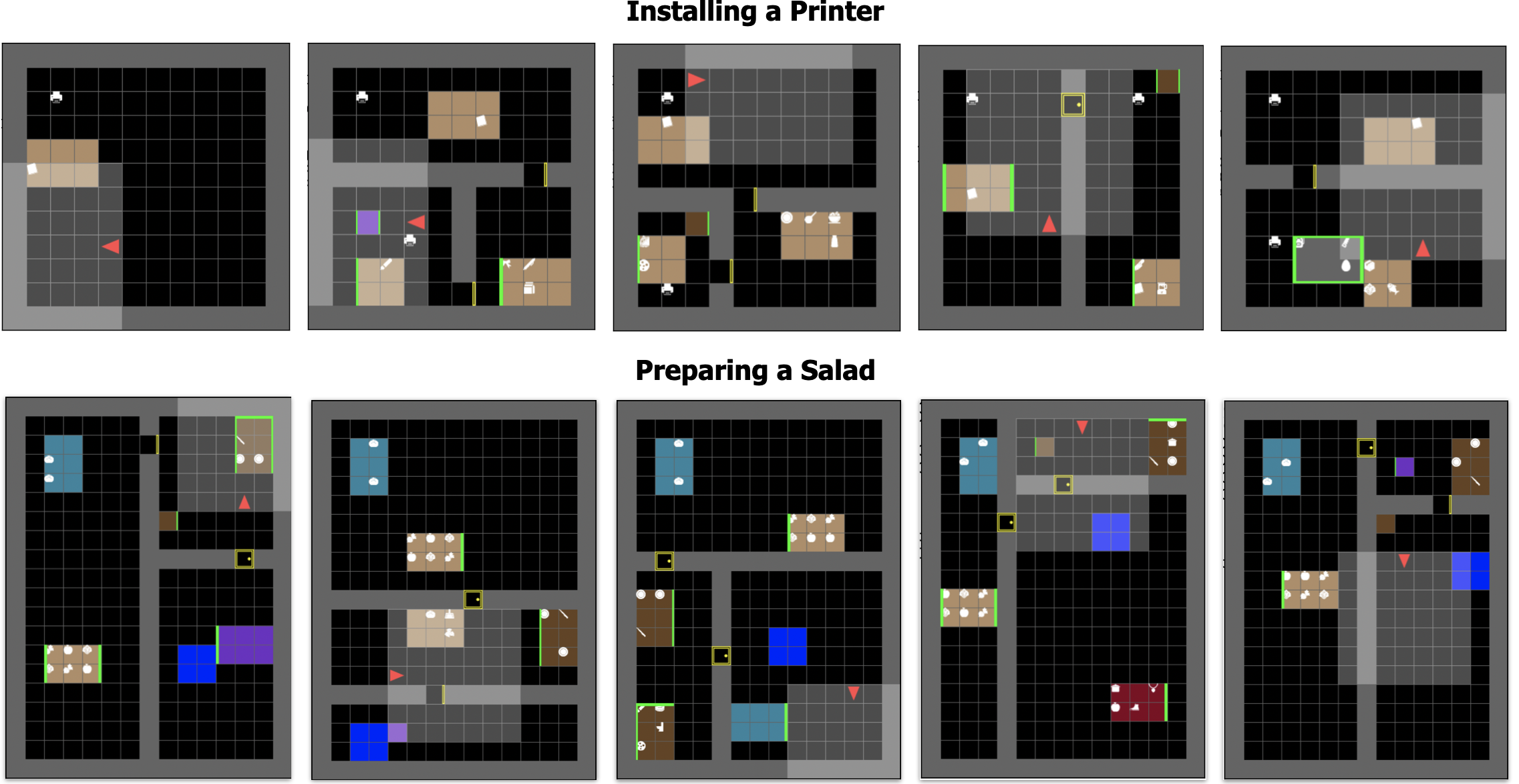}
    \caption{Examples of Procedural Generated Activity Instances: A few procedurally generated grid environments from procedural environment generation for two activities: \texttt{Installing a printer} and \texttt{Preparing a salad}.}
    \label{fig:pro_exp}
\end{figure}

\subsection{Diverse and Complex Tasks: For Reasoning and High-Level Planning}
In order to ensure that embodied AI agents learn the decision-making skills needed to complete real-world activities, Mini-BEHAVIOR tasks must reflect the wide range of difficulties and skills encountered in the real world.

Mini-BEHAVIOR provides a standardized set of 20 tasks chosen from BEHAVIOR-100 tasks. The chosen tasks represent a realistic distribution of diverse and complex household activities, including cleaning, organizing, and cooking tasks (see appendix A for a full list of activities and descriptions). The tasks are diverse in the sense that the activities involve a variety of household objects and require different states changes, such as moving objects, cleaning objects, and opening furniture. They are complex -- not only because they are long-horizon, but also since they require advancing reasoning skills for high-level planning.

Of all of the activities, the most simple one is \texttt{installing a printer}, in which the scene is initialized with a printer and table on the floor. To complete this, the agent must move the printer onto the table and toggle it on. This is simple since it involves interacting with only one object and relatively few number of high-level steps.

On the other hand, \texttt{washing pots and pans} is a hard activity that requires advanced high-level reasoning skills. The objective is to clean the stained kitchen appliances (teapot, kettle, pans) on the kitchen countertop and place them in the cabinets. To successfully accomplish this activity, the agent must execute a series of steps: navigate to find the relevant objects, pick up an unsoaked scrub brush, turn on the sink, soak the scrub brush in sink water, pick up the soap, clean each appliance using the soaked scrub brush and soap, and finally, place all of the items in the cabinet. This complex task requires the agent to interact with over 10 different objects and perform hundreds of individual actions in the correct sequence. This requires high-level planning and makes it difficult for the agent to complete.

By providing a standardized set of diverse and complex tasks with ranging difficulty, we hope that Mini-BEHAVIOR will facilitate more reasoning and high-level planning capabilities for embodied AI agents.

\subsection{Fast Simulation Environment: For Ease of Use and Rapid Prototyping}
Mini-BEHAVIOR contains a novel Gridworld environment with the capabilities to support diverse and complex tasks, yet is also designed to be lightweight and have minimal dependencies. The Mini-BEHAVIOR environment is built on top of MiniGrid, and it inherits its design philosophy for easiness of understanding and customization. 

It also has a comparably high simulation speed that makes it suitable for fast prototyping. On a regular laptop computer with Intel i7 eight-core processor at 1.90GHz, Mini-BEHAVIOR (without parallelization) can run at around 600 FPS.


\subsection{Procedural Generation: For Open-Ended Learning}

\begin{figure}[t!]
    \centering \includegraphics[width=\textwidth]{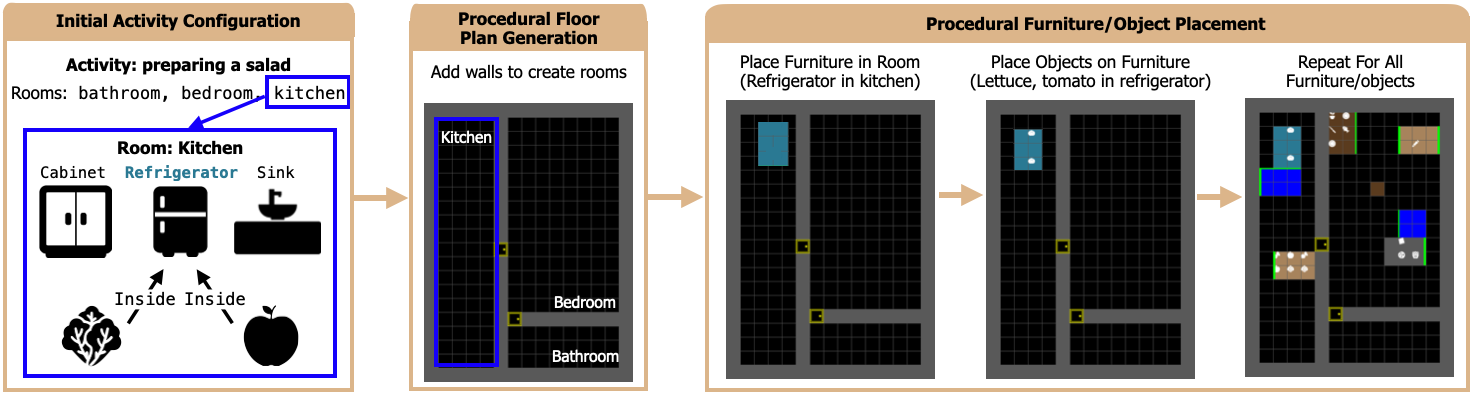}
    \caption{Procedural Generation Flow Chart: We allow procedural environment generation by simply taking in an initial activity configuration, with specifications for the grid environment, rooms, furnitures and objects. First, we generate a grid with a specified width and height, and then we procedurally generate a floor plan by adding walls and doors to create rooms. Then, we procedurally place the furnitures and objects: we add furniture to the specified room in the initial configuration, and then we place objects on the corresponding furniture with the correct symbolic states.}
    \label{fig:pro_flow}
\end{figure}

In the real-world, agents constantly encounter previously unseen environments, and they must be able to navigate and accomplish tasks under these novel settings. Mini-BEHAVIOR aims to support such open-ended learning by using procedural generation to provide an unlimited number of diverse activity settings. Inspired by ProcTHOR~\citep{procthor}, our procedural generation is achieved through procedural floor plan generation and procedural furniture/object placement, as shown in Fig.~\ref{fig:pro_flow}.

\textbf{Procedural Floor Plan Generation.}
The floor plan of the Gridworld household environment is determined by the layout and arrangement of rooms, walls, and doors. Mini-BEHAVIOR allows procedural floor plan generation by first generating the surrounding walls of the grid, randomly generating a wall with a door to split the grid into two connected rooms, choosing a room, and then repeating the wall generation process to further split the grid until a desired number of rooms is reached.


\textbf{Procedural Furniture/Object Placement.} The number and types of furniture and objects in the environment are predefined for the provided activities in their configuration files. 
Alternatively, they can be randomly sampled, where the number of furniture in each room will fall in the range $[1, \mathrm{max}(2, \frac{room\_width \times room\_height}{12}))$ and the number of objects on each furniture will fall in the range $[1, furniture\_width \times furniture\_height)$.
Given the number and types of furniture in each room defined, the pieces of furniture are randomly placed in the rooms one by one without space overlap. Similarly, with the number and types of objects on each piece of furniture defined, the objects are randomly placed on the corresponding furniture without space overlap. We also introduce a reachability check to make sure the floorplan and placements are valid by checking if all empty tiles are all connected.

This results in a procedurally generated activity instance that aids in open-ended learning, with examples shown in Fig.~\ref{fig:pro_exp}. 
\section{Mini-BEHAVIOR Environment Design}
The Mini-BEHAVIOR environment is designed to balance simplicity and speed while accommodating the complexity required to effectively support a wide range of long-horizon tasks. In addition to incorporating procedural generation, we extend beyond MiniGrid's capabilities to encompass symbolic states, an additional vertical dimension, multi-cell objects, and an abundance of actions, and objects. These enhancements introduce a level of realism essential for tackling complex tasks. Moreover, we incorporate various visual and control modes to enable different learning approaches -- such as visual agents and learning from human demonstrations.

\subsection{Motivation of Environment Features: Activity Definitions and Symbolic States}
To motivate the needs of our environment, it is important to understand how activities are defined and the role of symbolic states. Following BEHAVIOR, we define Mini-BEHAVIOR activities using the Behavior Domain Definition Language (BDDL), which is a predicate-based logic language that allows for infinite activity instances and flexible goal states. 

In BDDL, an activity definition consists of a set of objects, initial conditions, and goal conditions. Given an activity, Mini-BEHAVIOR procedurally generates environment instances with objects satisfying the initial conditions, and the agent completes the task by navigating and interacting with the environment until it fulfills all of the goal conditions. 

For example, the simplest activity of \texttt{Installing a Printer} is defined by the initial conditions \texttt{OnTop(printer, floor)}, \texttt{not ToggledOn(printer)}, \texttt{InRoom(table, office)}, and the task is considered complete when \texttt{OnTop(printer, table)} and \texttt{ToggledOn(printer)}. 


At any point in time, we use these symbolic states to describe the scene based on the physical objects' states (i.e. \texttt{ToggledOn}), their relationships to each other (i.e. \texttt{OnTop}), and their relationships to the agent (i.e. \texttt{InHandOf}). They are essential to our activity definitions and scene representations, so we design our environment to support these states both symbolically and visually (see appendix B for a list of all symbolic states and descriptions). 

We introduce three novel features to support symbolic states realistically -- a third environment dimension, multi-cell objects, and more comprehensive primitive actions. 

\subsection{A Novel Gridworld Environment}
In order to realistically simulate complex tasks and the physical relationships between objects, Mini-BEHAVIOR is a Gridworld environment that is designed to support vertical relationships between objects and notions of height.
Each Mini-BEHAVIOR environment is a 3D world with x-y-z axes and $n \times m \times 3$ cells. The z-axis represents a vertical dimension (top, middle, bottom) and introduces a notion of height and vertical placement to the environment. This is necessary to support more binary symbolic states and relations (i.e. on top, under, inside). 
Inspired by MiniGrid, each cell can contain at most one object. The limitation of 3 z-axis coordinates is still not entirely reflective of the real world, but choose this limitation to balance simplicity and realism in our environment. It is enough to support all of the BEHAVIOR tasks and, we believe, most real-world scenarios as well.

\subsection{Multi-Cell Objects}
We also introduce multi-cell furniture objects to reflect notions of size, which is important for physically representing the real world. Mini-BEHAVIOR supports two classes of objects: an object class and a furniture class. To reflect the real world, \textit{objects} are small objects that we often move or pick up in our everyday lives, such as a book or apple. On the other hand, \textit{furniture} are larger and stationary, and they include common household furniture such as a table, bed, and stove. In Mini-BEHAVIOR, the two distinctions between \textit{objects} and \textit{furniture} are that 1) \textit{objects} are $1\times1\times1$ in size while \textit{furniture} can span multiple cells, and 2) \textit{objects} are movable while \textit{furniture} are not.

\begin{figure}[t!]
    \centering    \includegraphics[width=1\textwidth]{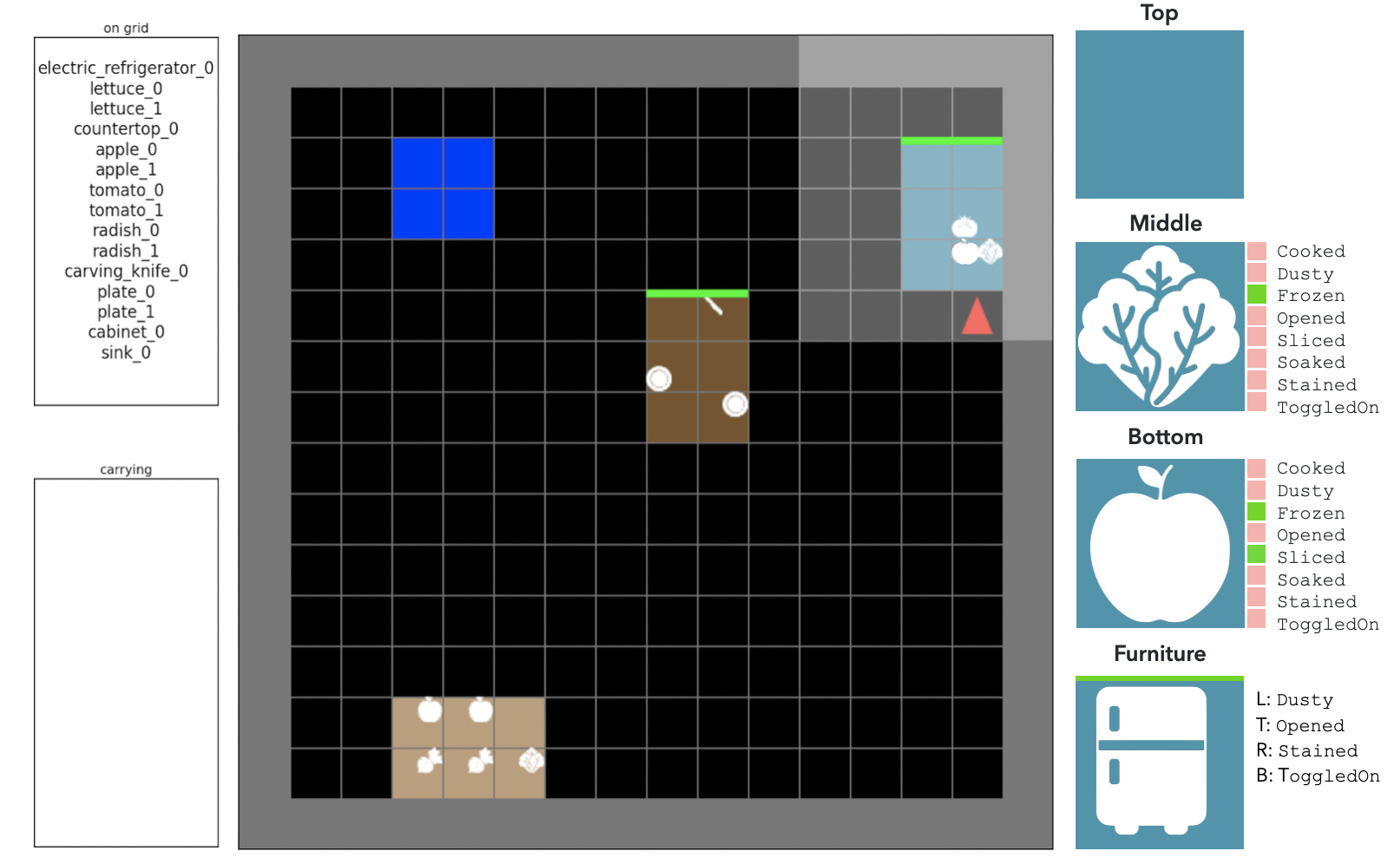}
    \caption{Visualization of the Mini-BEHAVIOR Environment: showing all objects as icons and all furniture as a colored background. All furniture states are shown in the grid with a green edge, while object states are only visualized for the cell directly in front of the agent by a red or green box.}
    \label{fig:default_view}
\end{figure}

\subsection{Agent}
\textbf{Action Space.} By default, Mini-BEHAVIOR agents use an action space that consists of 15 primitive actions -- 3 for navigation (forward, turn left, turn right) and 12 primitive actions for interacting with objects (close, cook, drop (3D), drop-in, open, pickup (3D), slice, toggle). This results in an 15-dimensional discrete action space. Under this setting, the same action space is shared across different tasks, allowing the study of transfer learning, multi-task learning, and meta-learning. 

A drawback of the primitive action space is that the agent can only carry one object at a time to prevent action ambiguity, which may cause inefficiency. Therefore, Mini-BEHAVIOR supports an additional type of action space - the Cartesian action space - which allows for the simultaneous carrying of multiple objects. The Cartesian action space consists of the Cartesian product between the primitive actions and available objects for a given task, with invalid actions (e.g. "slice a refrigerator") removed. Such an action space makes carrying and dropping multiple objects possible, and is often better suited for studying how to solve a single task. However, the Cartesian action space for different tasks will be different, based on the available objects. For a task where each object $i$ has $n_i$ possible actions, the action dimension of the Cartesian action space is equal to $4 + \sum_{i} n_i$. This ranges from 5 (for the simplest task of installing a printer) to 54 (for preparing a salad).

\textbf{Observation Space}. Mini-BEHAVIOR supports both fully observable and partially observable environments. In the partially observable case, the observation space of the agent is a $7 \times 7 \times 31$ ego-centric grid; when fully observable, the observation space is a $n \times n \times 31$ global grid, where $n$ is the size of the map. Each 31-dimension pixel encodes the states of the furniture and up to three objects that occupy a given cell on the map.

\textbf{Reward}. By default, Mini-BEHAVIOR uses sparse rewards for all tasks, where the agent receives +1 for successfully completing the task, and 0 otherwise.
We provide APIs such that the user can easily implement dense rewards for each environment, and toggle it on / off during environment initialization, exemplified in the following two tasks: putting away dishes after cleaning, and washing pots and pans.

\subsection{Additional Features to Support Research Needs}
Mini-BEHAVIOR is easy to use and customizable to fit the needs of researchers, and we also provide different visualization modes and agent control modes to support a variety of learning methods such as from visual input and human demonstrations. 

\textbf{Customizability}. By default, the layout of the environment is a $20\times20$ grid with walls along the outside. However, the layout is easily customizable. There are parameters which can be used to set the size and create grids of rooms, and we also provide some default layouts that correspond to scene layouts from iGibson, with examples shown in the Appendix. In addition, it is easy to create new tasks, objects, and actions using the classes and modules that we provide.

\textbf{Simulation Visualization}. To support visual agents and manual control of the environment, we provide various ways to visualize the environment as a grid rendering of $n\times m$ cells. The agent is always represented by a red triangle, and walls are represented by gray cells. Objects are represented by a corresponding object icon that is easy to recognize, and states are represented by a colored square. To visualize a 3D environment in a 2D format, we provide different views of the environment to provide a full view of the 3D environment. For example, the default view provides a birds-eye view of all objects in the environment (Fig~\ref{fig:default_view}), and a single-dimension view visualizes all objects in a single dimension and their unary states. We provide further discussion in the Appendix.

\textbf{Control Modes}
\label{sec:usage}
We support two modes: agent control and manual control. In the agent control mode, the agent is controlled by an algorithm, and this is useful for training different RL and planning agents. In manual control, a user can manually navigate and interact with the environment using keyboard controls. This is useful for creating human demonstrations.
 \section{Example Benchmark Results}
Mini-BEHAVIOR provides APIs both for training an agent with reinforcement learning and for collecting demonstration data that can be used for imitation learning. In this section, we provide preliminary results of training agents using proximal policy optimization (PPO) \cite{schulman2017proximal} on 3 different tasks -- \texttt{installing a printer}, \texttt{putting away dishes after cleaning}, and \texttt{washing pots and pans}. Based on the optimal number of steps to complete these tasks, we classify the difficulty of these tasks as easy, medium, and hard, respectively.
For all of these tasks, we use a grid size of $10\times10$, with a time limit of $1000$ steps per episode and a total of 1e6 training steps.  

\begin{figure}[t!]
    \centering
    \begin{subfigure}[b]{0.32\textwidth}
        \includegraphics[width=1.00\textwidth]{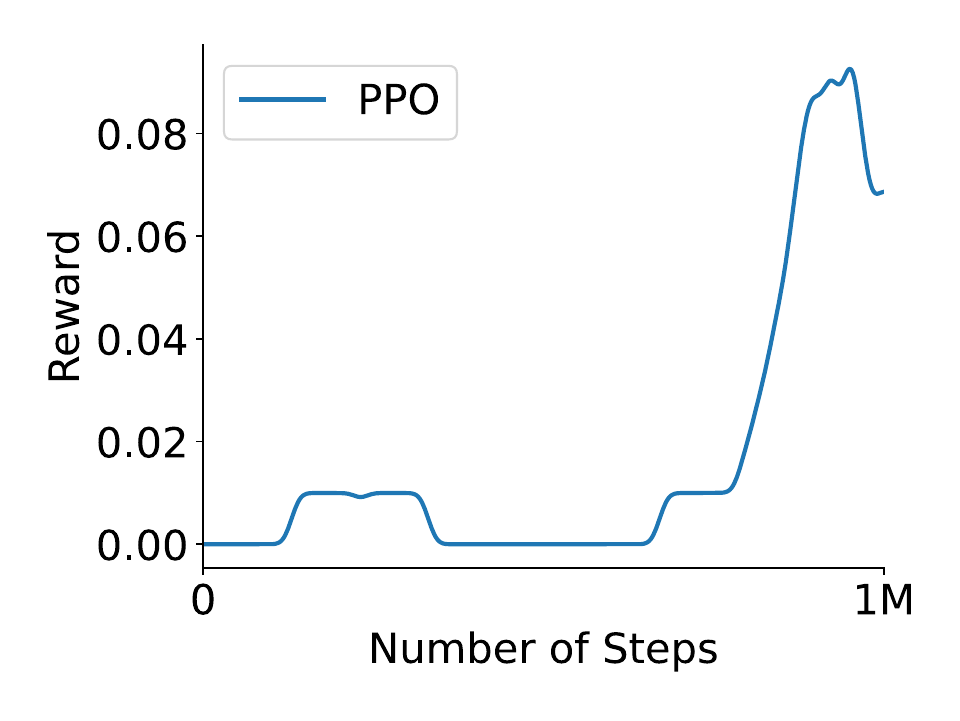}
        \caption{Printer (Sparse)}
        \label{fig:print}
    \end{subfigure}
    \begin{subfigure}[b]{0.32\textwidth}
        \includegraphics[width=1.00\textwidth]{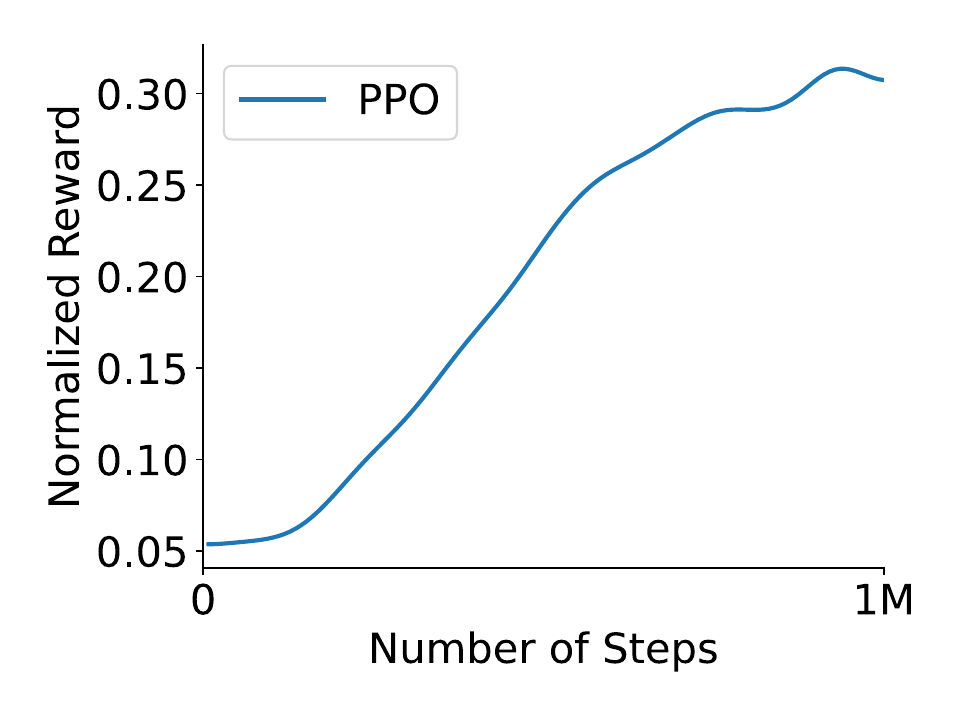}
        \caption{Dishes (Dense)}
        \label{fig:dish}
    \end{subfigure}
    \begin{subfigure}[b]{0.32\textwidth}
        \includegraphics[width=1.00\textwidth]{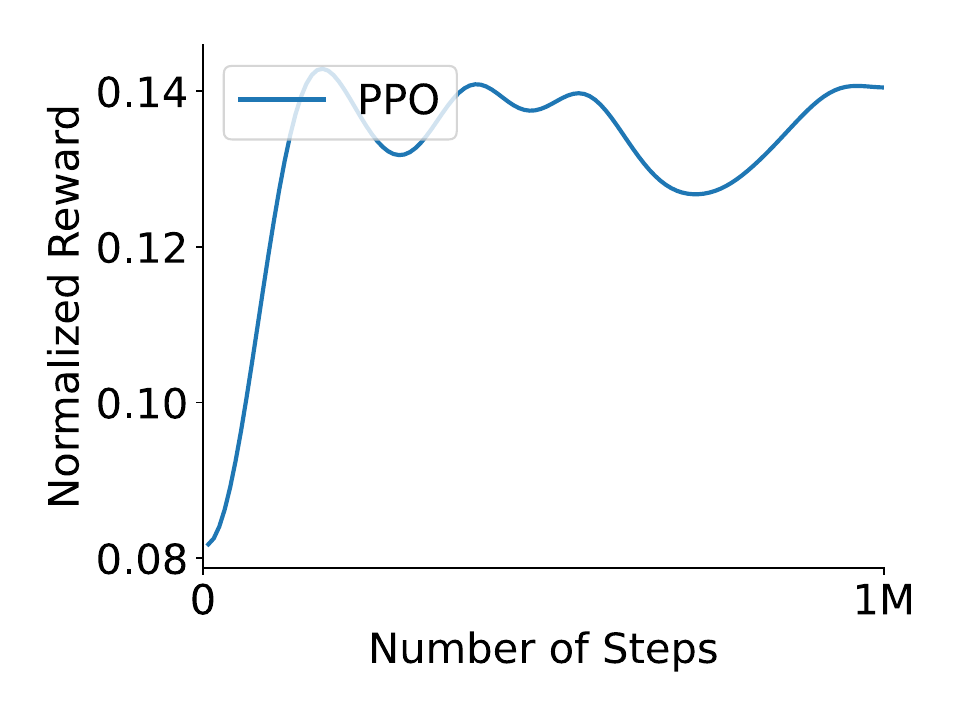}
        \caption{PotWash (Dense)}
        \label{fig:potwash}
    \end{subfigure}
            
    \caption{Preliminary results on the Mini-BEHAVIOR benchmark. We show results of vanilla PPO across three tasks, ranging from easy to hard. While simple and lightweight, Mini-BEHAVIOR is very challenging for cutting-edge decision-making algorithms, especially when a dense reward signal is not available.}
    \label{fig:rslts}
\end{figure}

We first examine these tasks under the sparse reward setting, where the agent will only receive a reward of +1 if it successfully completes the task. We found that, under the sparse reward setting, within 1M steps, \texttt{installing a printer} (the simplest out of the three) is the only task where the agent can achieve occasional success. We show the reward curve for \texttt{installing a printer} in Fig.~\ref{fig:print}, and omit the curves for the other two tasks due to them being zero throughout the training.

To gain further insights into the difficulty of the two harder tasks: \texttt{putting away dishes after cleaning} and \texttt{washing pots and pans}, we show additional results by hand-crafting a dense reward function for each of them. With dense reward, the agent receives a +1 reward every time it makes progress towards achieving the final goal (e.g. for \texttt{putting away dishes}, the agent receives a reward for opening the cabinet). We show the reward curves for the dense reward setting in Fig.~\ref{fig:dish} and Fig.~\ref{fig:potwash}. Notice that we normalize the reward in the plots such that a reward of 1 would correspond to successfully completing the task. Even with the dense reward, it is still difficult for vanilla PPO to learn a good policy for the medium and hard tasks, where the medium-level task agent has made a moderate amount of progress while the hard-level task agent is stuck at a relatively early stage.

These results together show that Mini-BEHAVIOR, while lightweight in terms of dependencies and computation, still provides a considerable amount of challenges necessary to evaluate state-of-the-art algorithms in decision-making and embodied AI.

\section{Conclusion}
In this paper, we introduce Mini-BEHAVIOR, a novel 3D gridworld simulation environment and benchmark of tasks to advance and support the development of high-level decision-making algorithms for embodied AI. Mini-BH is built on top of MiniGrid to be simple, lightweight, and fast, and it includes additional features to support simulating a set of 20 long-horizon, human-centered tasks for benchmarking. Mini-BH has already been used in other research projects \cite{wang2023a, kurenkov2023modeling}, and we hope that the embodied AI community as a whole will find value in using Mini-BH to prototype and benchmark decision-making algorithms that can ultimately be used to solve tasks in a realistic, physical simulation like those in BEHAVIOR.


\bibliographystyle{plainnat}
\bibliography{ref}

\newpage
\appendix 
\section{Task Descriptions}
Here, we describe the 20 tasks in Mini-BEHAVIOR.
\begin{enumerate}
    \item \textbf{Boxing books up for storage:} There are books and a box on the floor, and the agent must put all of the books in the box.
    \item \textbf{Cleaning a car:} There is initially a dusty car and unsoaked rag, and the agent must use the rag, soap, and sink to clean the car. After, the agent must place the rag and soap to a bucket. 
    \item \textbf{Cleaning shoes:} There are 2 stained shoes and 2 dusty shoes in the bedroom, and the agent must clean them using a towel and sink in the bathroom. 
    \item \textbf{Cleaning up the kitchen only:} There is a messy kitchen with a dusty floor, dusty pan, dusty cabinet, stained plate, and stained blender. The agent must find the broom, rag, and sink and use them to clean all of the dusty and stained objects.
    \item \textbf{Collect misplaced items: }There are items such as a gym shoe, necklace, notebook, and sock, misplaced around the living room and dining room. The agent must navigate these rooms, find the items, and place them on the table.
    \item \textbf{Installing a printer:} There is a printer on the floor, and the agent must place it on the table and toggle it on.
    \item \textbf{Laying wood floors:} There is are pieces of plywood, a saw, and a hammer on the floor. The agent must lay all of the plywood in the kitchen next to each other. 
    \item \textbf{Making tea:} There is a kitchen with a teapot, teabag, lemon, and knife hidden in various places like the cabinet and refrigerator. The agent must find these items and make tea by slicing a lemon, placing the teapot on the stove, placing the teabag inside of the teapot, and turning on the stove.
    \item \textbf{Moving boxes to storage:} There are two cartons on the floor of the living room, and the agent must move these to the storage room and stack them on top of one another.
    \item \textbf{Opening packages:} There are two unopened packages on the floor, and the agent must open both of them.
    \item \textbf{Organizing file cabinet:} There are various office objects (markers, folders, and documents) that are scattered on different furniture (chair, table, cabinet) in a private office. The agent must find the objects and place the markers on the table and documents and folders in the cabinet.
    \item \textbf{Preparing salad:} There are different salad ingredients (lettuces, apples, tomatoes, radishes), a plate, and a carving knife in the kitchen, and they are initially on the countertop, in the cabinet, or in the refrigerator. The agent must find these items, use the carving knife to slice the apples and tomatoes, and then place all of the salad ingredients on top of the plate.
    \item \textbf{Putting away dishes after cleaning:} There are plates on the kitchen countertop, and the agent must place all of them in the kitchen cabinet.
    \item \textbf{Setting up candles:} There are candles in a box, and the agent must take them out and place them on the table.
    \item \textbf{Sorting books:} There are books and hardbacks on the table, and the agent must place these on top of each other on the shelf.
    \item \textbf{Storing food:} There are boxes of oatmeal, chips, bottles of olive oil, and jars of sugar on the countertop. The agent must place all of these food items in the cabinet.
    \item \textbf{Thawing frozen food:} There are various frozen foods (fish, dates, and olives) in the refrigerator. The agent must take all of them out of the refrigerator and place them in the sink.
    \item \textbf{Throwing away leftovers:} There are hamburgers on plates, which are on countertops in the kitchen. The agent must throw all of the hamburgers into the trash can.
    \item \textbf{Washing pots and pans}: There are stained kitchen appliances (teapot, kettle, pans) on the kitchen countertop. The agent must clean the appliances using the scrub brush, soap, and sink and then place them in the cabinet.
    \item \textbf{Watering houseplants:} There are potted plants, and the agent must water them using the sink in the bathroom.
\end{enumerate}

\section{Symbolic States}
Mini-BEHAVIOR supports 3 types of symbolic states: agent-related, absolute, and relative object states:
\begin{itemize}
    \item \textbf{absolute object state}: a state that depends on a single object (i.e. \texttt{Cooked} and \texttt{ToggledOn})
    \item \textbf{agent-related state}: a state that represent relationships between the agent and object (i.e. \texttt{InHand} and \texttt{InSameRoom})
    \item \textbf{relative object state}: a state based on the physical relationships between two objects (i.e. \texttt{NextTo} and \texttt{OnTop}) 
\end{itemize}
This table lists the name, type, and description for all supported symbolic states:
\begin{center}
\begin{tabular}{ |c|c|c| } 
 \hline
 \textbf{Symbolic State} & \textbf{Type} & \textbf{Description} \\ 
 \hline 
 InFOV & agent & object is in the cell in front of the agent \\
  \hline
 InHand & agent & object is in the hand of the agent \\ 
 \hline
 InReach & agent & object is either in front or in hand of agent \\ 
 \hline
 InSameRoom & agent & object is in the same room as the agent \\ 
 \hline
 Cooked & absolute & the object is cooked (only food objects) \\
  \hline
 Dusty & absolute & the object is dusty \\ 
 \hline
 Frozen & absolute & the object is frozen (only food objects)  \\ 
 \hline
 Opened & absolute & the object is open (refregirator, box, etc) \\ 
 \hline
 Sliced & absolute & the object is sliced (only food objects)  \\ 
 \hline
 Soaked & absolute & the object is soaked \\  
 \hline
 Stained & absolute & the object is stained \\
 \hline
 ToggledOn & absolute & the object is on \\  
 \hline
 OnFloor & absolute & object is on the floor \\ 
 \hline 
 AtSameLocation & relative & object1 and object2 are in the same cell \\ 
 \hline
 Inside & relative & object1 is inside object2 \\ 
 \hline
 NextTo & relative & object1 and object2 are in adjacent cells \\ 
 \hline 
 OnTop & relative & object1 is one dimension above object2 \\ 
 \hline
 Under & relative & object1 is one dimension below object2 \\ 
 \hline
\end{tabular}
\end{center}  

\section{Room Size Customization}
\begin{figure}[h]
    \centering
    \includegraphics[width=0.9\textwidth]{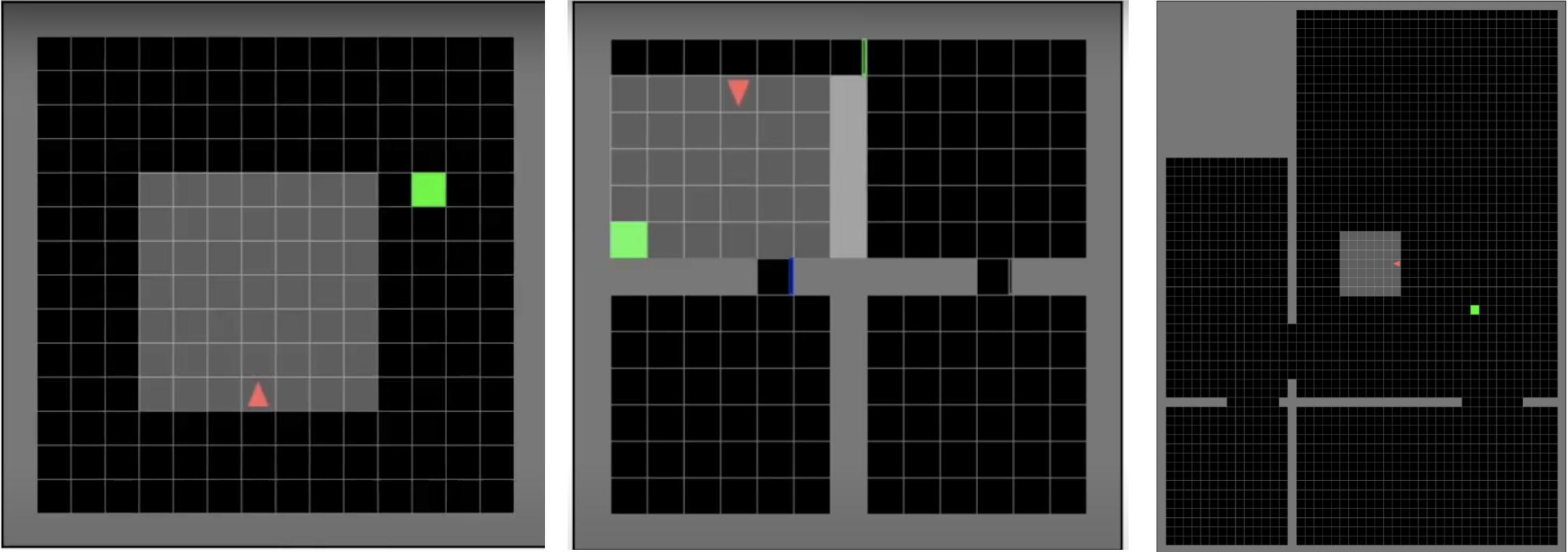}
    \caption{Examples of Mini-BEHAVIOR environment layouts: one $20\times20$ room (left), four $6\times6$ rooms (middle), iGibson \texttt{Rs\_int} scene layout (right)}
    \label{fig:env_layouts}
\end{figure}

Mini-BEHAVIOR allows for customization of environment layout size. In Fig.~\ref{fig:env_layouts}, we show examples of rooms with customized layout, including one which corresponds to the scene layouts from iGibson.

\section{Visualizations}

\subsection{State Visualization}
Only unary states are visualized, which means that the possible states are cooked, dusty, frozen, opened, sliced, soaked, stained, and toggledon for \textit{objects} (8 total) and dusty, opened, stained, toggledon for \textit{furniture} (4 total). 

To visualize the \textit{object} states, there are 8 squares to the right edge of the object cell to represent the 8 possible states. The square is green if the state is True and red otherwise. 

For \textit{furniture} states, note that all furniture is constrained to be rectangular, so we use the 4 edges to visualize the states (left: dusty, top: opened, right: stained, bottom: toggledon). In particular, a green edge indicates that a state is True.

\subsection{Scene Visualization}
We provide three ways to view the grid, each with its own characteristics and trade-offs.

\textbf{Default View}:
 By default, the scene visualization is a birds-eye-view of the environment that shows all of the objects in all dimensions (Fig.~\ref{fig:default_view}). \textit{Objects} are visualized as a simple yet easily recognizable icon, and are placed in different areas of the cell depending on the vertical dimension that it is in. \textit{Furniture} is rendered as a colored background. In this view, \textit{object} states are not shown, but \textit{furniture} states are. Still, it is possible to view the states of all objects in the cell directly in front of the agent as this includes all of the objects that the agent can interact with at that point. The user is able to open a 'Closeup' of the cell on the right of the grid, which displays 4 squares for the three dimensions and furniture in the cell. For each square, the icon and states are rendered. 

\begin{figure}[ht]
    \centering    \includegraphics[width=0.9\textwidth]{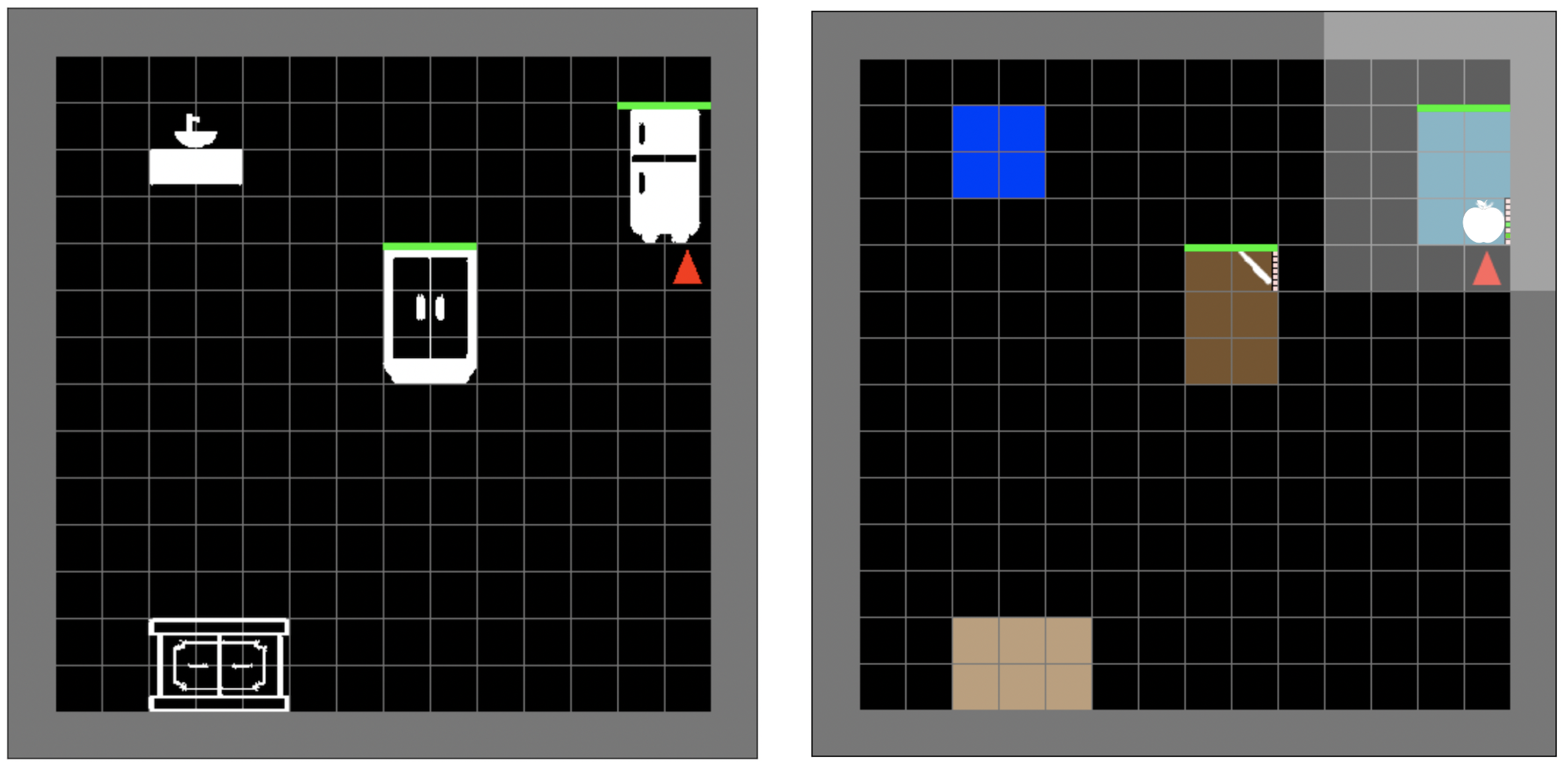}
    \caption{Other views of the grid -- (left) Furniture only view only shows furniture and furniture state. (right) Single dimension view of the bottom dimension, which shows all furniture and objects in the bottom dimension with their states. Comparing with the default view (Figure \ref{fig:default_view}), we see that the 'apple' from the closeup corresponds to the cell in front of the agent in the single dimension view}
    \label{fig:moreview}
\end{figure}

\textbf{Furniture Only}:
The dimension view (Fig~\ref{fig:moreview}, right) is used to visualize a single vertical dimension (top, middle, or bottom), and it shows all objects, furniture, and states in the specified dimension. Similar to the default view, objects are rendered as icons while furniture is a colored background. The key difference is that the object icon takes up the full cell, and the object states are visualized in addition to the furniture states.

\textbf{Single Dimension}:
In the furniture-only view (Fig~\ref{fig:moreview}, left), the \textit{furniture} icons are displayed, and no \textit{objects} or object states are shown. 

\end{document}